# False perspectives on human language: why statistics needs linguistics

Matteo Greco[1]*, Andrea Cometa[1,2], Fiorenzo Artoni[3,4], Robert Frank[5], Andrea Moro[1]

**Affiliations:**

[1] University School for Advanced Studies IUSS Pavia; Pavia, 27100, Italy.

[2] The Biorobotics Institute and Department of Excellence in AI and Robotics, Scuola Superiore Sant'Anna, Pisa, Italy.

[3] Ecole Polytechnique Federale de Lausanne, Bertarelli Foundation Chair in Translational NeuroEngineering, Center for Neuroprosthetics and School of Engineering, Chemin des Mines 9, Geneva, GE CH 1202, Switzerland

[4] Functional Brain Mapping Laboratory, Department of Basic Neurosciences, University of Geneva, Chemin des Mines 9, Geneva, GE CH 1202, Switzerland

[5] Department of Linguistics, Yale University; New Haven, United States of America.

*Corresponding author. Email: matteo.greco@iusspavia.it

**Abstract:** A sharp tension exists about the nature of human language between two opposite parties: those who believe that statistical surface distributions, in particular using measures like surprisal, provide a better understanding of language processing, vs. those who believe that discrete hierarchical structures implementing linguistic information such as syntactic ones are a better tool. In this paper, we show that this dichotomy is a false one. Relying on the fact that statistical measures can be defined on the basis of either structural or non-structural models, we provide empirical evidence that only models of surprisal that reflect syntactic structure are able to account for language regularities.

**One-Sentence Summary:** Language processing does not only rely on some statistical surface distributions, but it needs to be integrated with syntactic information.

A sharp tension exists about the nature of human language between two opposite parties: those who believe that statistical surface distributions, in particular characterized using measure like surprisal, provide a better understanding of language processing, vs. those who believe that discrete recursive hierarchical structures implementing linguistic information are a better tool, more specifically, syntactic structures, the core and unique characteristic of human language (7). In this paper, we show that this dichotomy is a false one. Relying on the fact that statistical measures can be defined on the basis of either structural or non-structural models, we provide empirical evidence that only models of surprisal that reflect syntactic structure are able to account for language regularities.

**1. On four different models of surprisal**

It is a truism that during language processing the brain computes expectations about what material is likely to arise in a given context. The natural next step from this observation and one that characterizes much work in psycholinguistics is to formulate a hypothesis about the differences in processing load: in general, the less expected a piece of linguistic material is, the more difficult its processing (8, 13). Expectation can be quantified in terms of the information theoretic notion of Surprisal (2), where the surprisal of a word $w$ in context $w_c$ is defined as:

$$Surprisal(w|w_c) = -\log p(w|w_c) \quad (1)$$

If a word is highly unlikely in a context, its surprisal will be very high. In contrast, if the word's is highly likely, its surprisal will approach 0.

Surprisal serves as a very useful linking hypothesis between patterns of behavior and brain response on the one hand and a single numerical quantity, namely the probability of a form. And because surprisal does not make explicit reference to linguistic structure, surprisal is often thought to provide an alternative perspective on language processing that avoids the necessity to posit such structure. This view is incorrect, however. Surprisal depends crucially on a particular characterization of a word's probability. Such a characterization, a probability model, may or may not make reference to linguistic structure. In this section, we will describe two dimensions along which language probability models can vary, and then use these dimensions to characterize four distinct probability models. Each of these models can be used as the input to the surprisal equation given above, so that different values of surprisal can result depending on the assumptions behind the probability model (see Fig. 1).

*1.1 Dimension 1: Sequences vs. Hierarchical Structure*

Our first dimension concerns the structure that is assumed in the generation of language. The simplest conception views language as a concatenative system. In this view, a sentence is simply a sequence of words generated one after another in a linear fashion. To account for which sentences are well-formed and which are not, constraints are imposed on adjacent elements, or bigrams. For example, in the context preceded by word 'the', a linear model of English will permit words like 'cat' or 'magazine' to occur, but not 'of'. To make a probability language model, we can simply assign a probability to a word $w$ in a given context defined by the previous word $w_c$, so that the probabilities for all of the words sum to 1 for each context. Given a sufficiently large corpus, we can estimate these probabilities by taking the ratio of the number of occurrences of the context and of the context-word bigram:

$$p(w|w_c) = \frac{count(w_c, w)}{count(w_c)} \quad (2)$$

This model can be extended to an n-gram model, where the length of the context is increased to include more material: in an n-gram model, the conditioning context will include n-1 words. A 3-gram model could thus assign a higher probability to 'magazine' than 'cat' in the context 'read the' while doing the reverse in the context 'fed the'. A bigram model could not assign distinct probabilities in the two contexts, since the single adjacent word, namely 'the', is identical in both. For this reason, an n-gram model gives a more refined assessment of likelihood as the value of n grows. However, because the number of conditioning contexts expands exponentially with the length of the context, it becomes increasingly difficult to accurately estimate the values of the probability model. A variety of methods have been proposed to integrate the information from longer contexts with information in shorter contexts. We use such a composite model for our model of **N-gram surprisal**.

Chomsky (1957) (4) famously argued that linear models, were inadequate models of natural language, as they are incapable of capturing unbounded dependencies. To illustrate, consider the likelihood of the word 'is' or 'are' in context 'The book/books that I was telling you about last week during our visit to the zoo'. This will depend on the whether the word 'book' or 'books' appears in the context. Because the distance between this contextual word and the predicted verb can grow without bound, no specific value of n will yield an n-gram model that can correctly assign probability in such cases.

Chomsky's suggested alternative generates language using a hierarchically organized process. In this way, linearly distant elements can be structural close. One simple model for this involves context-free grammars (CFG), a set of rules that specify how a unit in a sentence tree can be expanded:

S → NP VP

NP → Det N

VP → V NP | V

Det → the | a

N → book | books

V → read | reads

Where S is the sentence, NP is a noun phrase, VP is a verb phrase, Det is a determiner, N is a noun and V is a verb.

Generating a sentence with such a grammar starts at the start symbol S. A rule whose lefthand side matches this symbol is then selected to expand the symbol. Each element of this expansion is in turn expanded with an appropriately matching rule, until the only remaining unexpanded symbols are words. The result of this CFG derivation is a tree-structured object T, whose periphery consists of the words of the sentence that is generated, called the yield of T. A CFG can be used as the basis of a probability model by assigning probability distributions for the possible expansions of each symbol (i.e., a value between 0 and 1 is assigned to each rule, with the values for the rules that share the same lefthand side summing to 1). In such a probabilistic CFG (PCFG), derivations proceed as with CFGs, but the choice of expansions is determined by the probabilities. In PCFGs, the probability of a tree structure is the product of the probabilities of each of the expansions. Because a sequence of words S might be generated by different trees, the probability of S is the sum of the probabilities of all of the trees T with yield

S. Hale (2001) (9) shows how to use PCFGs to calculate the surprisal for a word given a context: we take the summed probability of all trees whose yield begins with the context-word (i.e., the prefix probability for context-word) divided by the summed probability of all trees whose yield begins with the context (i.e., the prefix probability for context).

PCFGs of this form suffer from being unable encode dependencies between lexical items: the choice of the verb in a VP is made independently of the choice of the noun in the verb's NP object. A body of work in the literature in natural language processing has addressed this shortcoming by adding 'lexicalization' to a PCFG, and this is the approach we adopt, following (11).

*1.2 Dimension 2: Word vs. Category prediction*

As already noted, n-gram models with longer contexts suffer from an estimation problem: it is impossible to get accurate estimates of the likelihood of relatively infrequent words in contexts that are defined by sequences of, say, 5 words. We can avoid this problem by incorporating another aspect of abstract linguistic structure: the categorization of words in part-of-speech (POS) classes. We can define a POS n-gram model as one where both the context (and the predicted element are POS (e.g., noun, verb, determiner, etc.). To compute the surprisal of a word w, then, equation (2) becomes:

$$p_{POS}(w|w_c) = \frac{count(c_c, c)}{count(c_c)} \quad (3)$$

where $c_c$ is the POS of the context, and $c$ is the POS of the target word.

This is what we use for our model of **POS surprisal**.

With a small set of POS labels, the probability values for longer n-grams can be accurately estimated. Note though that POS n-gram model is insensitive to the meaning of individual words, so it will be unable to distinguish the probability of 'cat' and 'magazine' occurring in any context, as they are both nouns, but could distinguish their likelihood from that of prepositions like 'of' or adjectives like 'furry'. As a result, this model's predictions for surprisal will differ from those of a word-based surprisal model.

Roark et al. (2009) (11) propose a method for separating between word vs. category prediction in the context of a hierarchy-sensitive probability models. Specifically, for the category predictions, the prefix probability of the context-word sequence omits from the probability of the generation of the word. Following Roark et al., we call the resulting surprisal predictions Syntactic Surprisal. For word predictions, on the other hand, the context includes not only that contributed by the preceding words, but also the structure up to, but not including, the generation of the word. Again following Roark et al. (11), we call the surprisal values computed in this way Lexical Surprisal.

Fig. 1

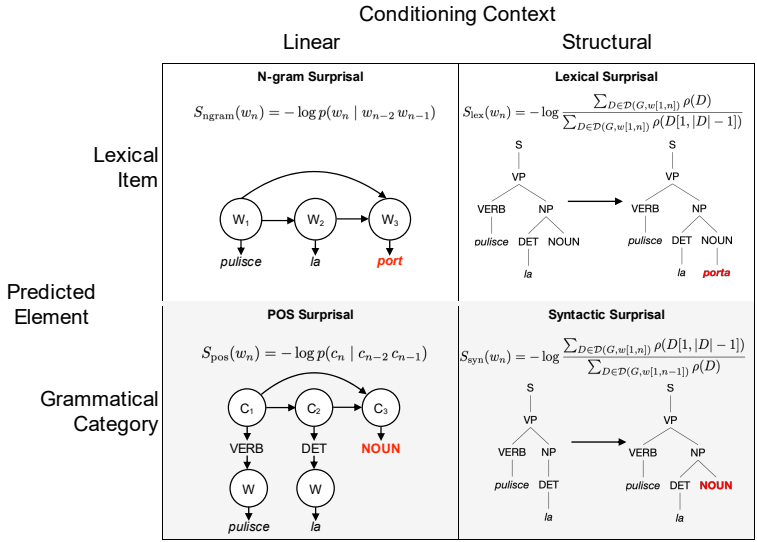

Fig. 1: The two dimensions of language models (linear vs. hierarchical structure and word vs. category prediction). A choice in each dimension yields a distinct model of language, from which we can extract probability values.

## 2. Challenging data

In order to test different types of surprisal models a new set of stimuli has been designed building on Artoni et al. (2020) (1). In that work the neural decoding of linguistic structures from the brain was found in carefully controlled data, where confounding factors such as acoustic information were factored out distinguishing this work from previous in the field such as in (3, 6, 12). Specifically, their stimuli involved pairs of sentences sharing strings of two words with exactly the same acoustics (*homophonous phrase*, hence HP) but with completely different syntax. This strategy was made possible by relying on the properties of the Italian language. HPs could be either a Noun Phrase (NP) or a Verb Phrase (VP), depending on the syntactic structure that is involved. More specifically, HPs contained two words, such as *la porta* [laˈpɔrta]: a first monosyllabic word (e.g., *la*) which could be interpreted either as a definite article (Eng. 'the$_{\text{fem.sing.}}$') or an object clitic pronoun (Eng. 'her'); a second polysyllabic word (e.g., *porta*) which could be interpreted either as a noun (Eng. 'door'), or a verb (Eng. "brings"). The whole HP could be interpreted either as a NP ('the door') in *Pulisce **la porta** con l'acqua* (s/he cleans **the door** with water) or as a VP ('brings her') in *Domani **la porta** a casa* (tomorrow s/he brings her home) depending on the syntactic context within the sentence where they were pronounced. Unfortunately, it was not completely possible to disentangle surprisal from the syntactic information, since the linguistic material preceding HPs was different in the NPs vs. VPs interpretation, such as in ***Pulisce** la porta* (**s/he cleans** the door) vs. ***Domani** la porta* (**tomorrow** s/he brings). Relying on this kind of sentences, three experimental conditions have been generated here by modulating the syntactic context preceding HPs, which predicts the syntactic type of HPs, in order to modulate the relation between syntactic and surprisal information:

(i)     ***unpredictable* HPs** (UNPRED)**:** the syntactic context preceding HPs allows both NPs and VPs since it is an adverb. Therefore, the syntactic types of HPs are not predictable at the beginning of the sentence, but only after the HPs: if HPs are followed by verbs (such as in ***Forse** la porta è aperta*, 'Maybe the door is open') they realize NPs, otherwise they realize VPs (***Forse** la porta a casa*, 'Maybe s/he brings it at home').

Since the lexical context preceding HPs is exactly the same for both NPs and VPs, no differences in the surprisal value can be detected at the HP.

(ii) **Strong predictable HPs** (Strong_PRED): the syntactic context preceding HPs allows either NPs or VPs (but not both) and, therefore, the syntactic type of HP is predictable at the beginning of the sentence: if HPs are preceded by verbs, they realize NPs (such as in **Pulisce** *la porta con l'acqua*, 'S/he cleans the door with water'); if HPs are preceded by nouns, they realize VP (such as in **La donna** *la porta domani*, 'A woman brings her tomorrow'). This was the kind of stimuli exploited in Artoni et al. (2020), where the lexical context preceding HPs was different in NPs and VPs, allowing different surprisal values in the two cases.

(iii) **Weak predictable HPs** (Weak_PRED): the syntactic context preceding HPs allows both NPs and VPs, as in the *unpredictable* HPs, thus the first word of the HP (*la*) could either be an article or a clitic pronoun, but the second word of the HP (*porta*) can only be analyzed as a noun (door), as in 1$^{st}$ class predictable HPs, since the temporal adverb introducing the sentence (such as *ieri*, 'yesterday') requires a past tense whereas the verbal form of the HP displays a present tense (brings) (such as **Ieri** *la porta era aperta*, 'Yesterday the door/*brings it was open'). As in the unpredictable class, the surprisal value is eliminated by the lexicon preceding HPs, which is the same for both NPs and VPs (only the morphosyntactic shape of the second HP word forces the interpretation forward the NP).

A total of 150 trials were prepared: 60 for UNPRED-HPs, 30 UNPRED-NPs and 30 UNPRED-VPs, 60 for Strong_PRED-HPs, 30 Strong_PRED-NPs and 30 Strong_PRED-VPs, and 30 for Weak_PRED-HPs, only Weak_PRED-NPs since there cannot be VPs of this type.

## 3. Statistical Analysis

Here we compared the N-gram, Lexical, POS and Syntactic surprisal of the 5 classes of stimuli (Strong_PRED-NP, Strong_PRED-VP, Weak_PRED-NP, UNPRED-NP, UNPRED-VP) relative to the first and the second word of the HPs. Kruskal-Wallis tests revealed significant differences across the surprisal values associated with all five classes for all notions of surprisal. For the nouns and verbs, the difference was significant only for the POS surprisal and the syntactic surprisal. We further investigated these differences using Conover post-hoc tests with Holm-Bonferroni correction. For the articles and clitics, only the syntactic surprisal captured the difference across all three classes of predictable items ($p<0.0001$, **Fig 2,** top row). The POS and N-gram surprisal values of the articles were lower than those of the clitics ($p<0.05$), while the lexical surprisal values of the articles of the Strong_PRED-NP sentences were lower than the lexical surprisal values of the articles of weak_PRED-NP sentences and the clitics of Strong_PRED-VP sentences. For nouns and verbs, both the POS surprisal and the syntactic surprisal showed a difference between all three stimuli classes ($p<0.05$, **Fig 2,** bottom row). There was no difference between the N-gram surprisal values or lexical surprisal values of nouns and verbs.

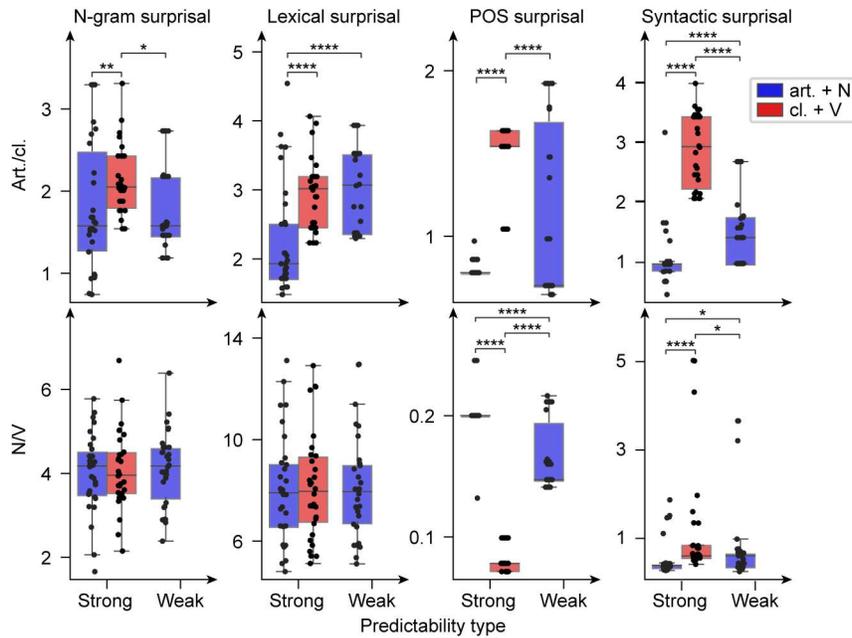

Fig 2. **Predictable items by class**. Boxplots of the surprisal values for the (strong and weak) predictable items for the articles/clitics (art./cl., top row) and the nouns/verbs (N/V, bottom row). Each column represents a distinct notion of surprisal. * $p < 0.05$, ** $p < 0.01$, *** $p < 0.001$, **** $p < 0.0001$.

For the unpredictable items, only the POS surprisal values were different between the articles and clitics and between the nouns and verbs (**Fig 3**).

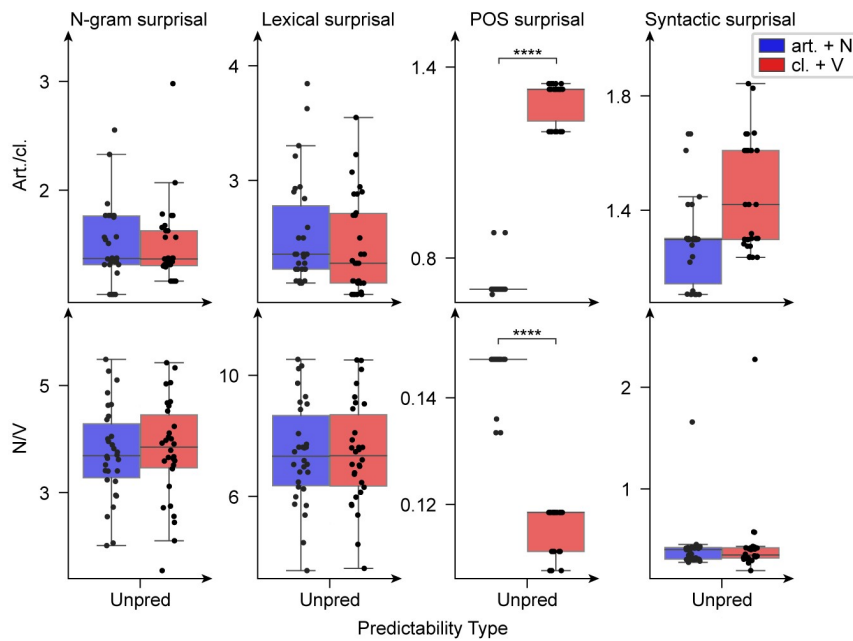

Fig 3. **Unpredictable items**. Boxplots of the surprisal values for the unpredictable items for the articles/clitics (art./cl., top row) and the nouns/verbs (N/V, bottom row). Each column represents a distinct notion of surprisal. * $p < 0.05$, ** $p < 0.01$, *** $p < 0.001$, **** $p < 0.0001$.

We defined four different classification tasks: Strong_PRED nouns vs. verbs (i), predictable (Strong_PRED and Weak_PRED) nouns vs. verbs (ii), UNPRED nouns vs. verbs (iii), and predictable items vs. unpredictable items (iv). For each classification task we trained and validated (10-fold cross validation) one Support Vector Machines (SVMs) for each notion of surprisal (i.e. using the values calculated according to the given notion of surprisal as features), and one SVM trained on all surprisal values regardless of the surprisal type, called tot-SVM. For classification tasks (i), (ii), and (iv), the SVMs trained on POS surprisal, Syntactic surprisal, and the tot-SVM reached near 100% accuracy, above the other two classifiers ($p<0.05$, Conover post-hoc with Holm-Bonferroni correction). For classification task (iii), tot-SVM and the POS surprisal-trained SVM reached 100% accuracy, while Syntactic surprisal-SVM achieved slightly above-chance accuracy (**Fig 4**).

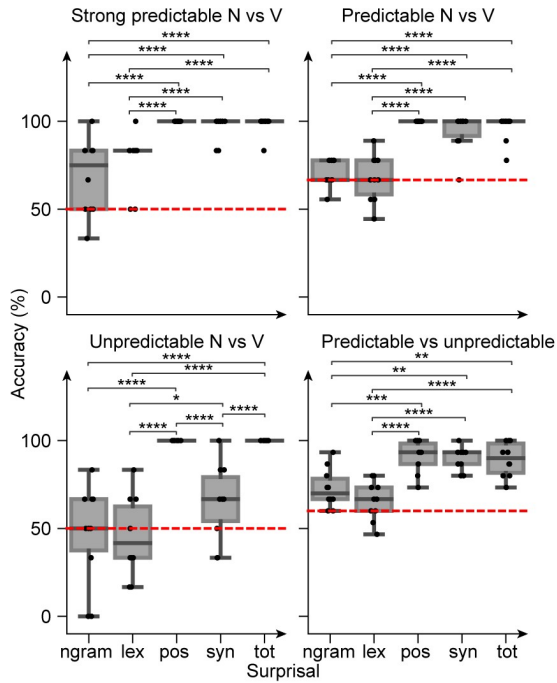

Fig. 4 **Decoding results**. Boxplots of the accuracies for the distinct classification tasks using different sets of features. Each data point is the accuracy of 1 fold in a 10-fold cross validation procedure. The red dashed lines are the chance levels. Strong Predictable N vs. V: classification task (i). (Strong and weak) Predictable N vs. V: classification task (ii). Unpredictable N vs. V: classification task (iii). Predictable vs. unpredictable: classification task (iv). For each set of features both the surprisal of the article/clitic and of the noun/verb were considered. The set of features are: ngram – N-gram surprisal; lex – Lexical surprisal; pos – POS surprisal; syn – Syntactic surprisal; tot – all of the above. * $p < 0.05$, ** $p < 0.01$, *** $p < 0.001$, **** $p < 0.0001$.

## 4. Discussion and conclusion.

In this paper four different probability models of surprisal have been compared by exploiting the following contrasting factors: words vs. parts-of-speech and sequences vs. hierarchical structures. In order to test these models three experimental conditions have been generated by modulating the surprisal context: those where the phrase was completely unpredictable by the contexts (unpredictable phrases), those where the phrase was immediately predictable by the first word of the phrase (strong predictable phrases), and those where the phrase was predictable only after the second word of the phrase (weak predictable phrases). Notably, all confounding factors, including acoustic information, were factored out distinguishing our work from previous in the field such as in (3, 6, 12). We found that only those models combining hierarchical structures and part-of-speech categories successfully fit the three classes. On the other hand, surprisal models that only considers sequences of both words and parts-of-speech fail to replicate the expectation associated to the three classes. More specifically, statistical surface distributions are proved to be largely insufficient when it comes to language structure.

**References and Notes.**

1. F. Artoni, P, d'Orio, E. Catricalà, et al. High gamma response tracks different syntactic structures in homophonous phrases. *Sci Rep* **10**, 7537 (2020). https://doi.org/10.1038/s41598-020-64375-9

2. F. Attneave, *Applications of Information Theory to Psychology: a summary of basic concepts, methods and results* (Rinehart and Winston, 1959).

3. J. R. Brennan, J. T. Hale, Hierarchical structure guides rapid linguistic predictions during naturalistic listening. *PloS one* **14 (1)**, e0207741 (2019).

4. N. Chomsky, *Syntactic Structures* (Mouton, 1957).

5. C. Cortes, V. Vapnik, Support-vector networks. *Mach. Learn*. **20**, 273–297 (1995).

6. S. L. Frank, L. J. Otten, G. Galli, G. Vigliocco. The ERP response to the amount of information conveyed by words in sentences. *Brain and Language* **140** 1-11 (2015).

7. A. Friederici, *Language in our brain: the origins of a uniquely human capacity* (MIT Press 2017).

8. F. Goldman-Eisler, Speech production and the predictability of words in context. *Quarterly Journal of Experimental Psychology* **10(2)**, 96–106 (1958).

9. J. Hale, "A probabilistic Earley parser as a psycholinguistic model" in *Proceedings of the Second Meeting of the North American Chapter of the Association for Computational Linguistics on Language Technologies (NAACL)*. Association for Computational Linguistics (2001). 10.3115/1073336.1073357.

10. F. Pedregosa, G. Varoquaux, A. Gramfort, V. Michel, B. Thirion, O. Grisel, M. Blondel, P. Prettenhofer, R. Weiss, V. Dubourg, J. Vanderplas, A. Passos, D. Cournapeau, M. Brucher, M. Perrot, É. Duchesnay, Scikit-learn: Machine Learning in Python. *J. Mach. Learn. Res.* **12**, 2825–2830 (2011).

11. B. Roark, A. Bachrach, C. Cardenas, C. Pallier, "Deriving lexical and syntactic expectation-based measures for psycholinguistic modeling via incremental top-down parsing" in *Proceedings of the 2009 Conference on Empirical Methods in Natural Language Processing: Volume 1 (EMNLP)* (Association for Computational Linguistics, 2009), pp. 324–333.

12. C. Shain, I. A. Blank, M. van Schijndel, W. Schuler, E. Fedorenko, fMRI reveals language-specific predictive coding during naturalistic sentence comprehension. *Neuropsychologia* **138** 107307 (2020).

13. W.L. Taylor, "Cloze procedure": a new tool for measuring readability. *Journalism quarterly* **30(4)**, 415–433 (1953).

**Acknowledgments:** The authors would like to thank Silvestro Micera, Stefano Cappa and Claudia Repetto as part of the working group of the PRIN-INSPECT project and for their precious suggestions.

**Funding:** Ministero dell'Università e della Ricerca (Italy) grant: INSPECT-PROT. 2017JPMW4F_003

**Author contributions:**

Conceptualization: AM.

Methodology: MG, AC, FA, RF.

Visualization: AC, RF.

Funding acquisition: AM

Supervision: MG, AC, FA, RF, AM.

Writing – original draft: MG, AC, FA, RF, AM.

Writing – review & editing: MG, AC, FA, RF, AM.

**Competing interests:** Authors declare that they have no competing interests.

**Data and materials availability:** the data that support the findings of this study are available upon reasonable request from the authors.

**Supplementary Materials**

Supplementary Text: Decoding of stimulus class

# Supplementary Materials for

**False perspectives on human language: why statistics needs syntax**

Matteo Greco[1]*, Andrea Cometa[1,2], Fiorenzo Artoni[3], Robert Frank[4], Andrea Moro[1]

Corresponding author: matteo.greco@iusspavia.it

**Supplementary Text**

<u>Decoding</u> <u>of</u> <u>stimulus</u> <u>class</u>

For each classification task, and for each notion of surprisal, a Support Vector Machine (SVM) (5) with a radial basis function kernel was trained. Each SVM received as input the surprisal values of both the art./cl. and the N/V calculated according to the corresponding method (i.e. two features for each SVM). The tot-SVM was trained on the combination of all surprisal values for the art./cl. and the N/V calculated according to all the four different notions of surprisal (i.e., 8 features).

The training was carried out using a nested cross-validation procedure: (i) a first 10-fold cross-validation was used to split the dataset into training and test set, and (ii) for each fold of the 10-fold cross-validation, another 10 fold cross-validation was used to furtherly divide the training set into training and validation set.

The inner validation loop was used to optimize the decoder hyperparameters. The optimized hyperparameters were: C, i.e. the cost of misclassification of training instances; and the free parameter of the radial basis function gamma. Hyperparameter optimization of was carried out using a grid search on [0.001, 0.01, 0.1, 1, 10] for C and [0.001, 0.01, 0.1, 1] for gamma. For each fold of the outer validation loop, the best hyperparameters were set as the C and gamma values which achieved the best mean accuracy in the inner 10-fold cross validation loop, thus resulting in a different set of hyperparameters for each fold of the outer validation loop.

Data points in **Fig 3** represent the accuracy values of each fold of the outer loop.

Code implementation was based on the scikit-learn package for python (10).